\documentclass[10pt, a4paper]{article}
\usepackage{stfloats}
\usepackage{lrec-coling2024}

\title{\textbf{Locations of Characters in Narratives:\\ Andersen and Persuasion Datasets}}

\name{Batuhan Özyurt, Roya Arkhmammadova, Deniz Yuret} 

\address{Koç University \\
         \{bozyurt20, rarkhmammadova22, dyuret\}@ku.edu.tr\\}

\abstract{
The ability of machines to grasp spatial understanding within narrative contexts is an intriguing aspect of reading comprehension that continues to be studied. Motivated by the goal to test the AI's competence in understanding the relationship between characters and their respective locations in narratives, we introduce two new datasets: Andersen and Persuasion. For the Andersen dataset, we selected fifteen children's stories from "Andersen's Fairy Tales" by Hans Christian Andersen and manually annotated the characters and their respective locations throughout each story. Similarly, for the Persuasion dataset, characters and their locations in the novel "Persuasion" by Jane Austen were also manually annotated. We used these datasets to prompt Large Language Models (LLMs). The prompts are created by extracting excerpts from the stories or the novel and combining them with a question asking the location of a character mentioned in that excerpt. Out of the five LLMs we tested, the best-performing one for the Andersen dataset accurately identified the location in 61.85\% of the examples, while for the Persuasion dataset, the best-performing one did so in 56.06\% of the cases.
 \\ \newline  }

\usepackage{listings}
\usepackage{color}
\usepackage{float}

\usepackage{hyperref}
\usepackage{url}
\usepackage{xurl}

\definecolor{dkgreen}{rgb}{0,0.6,0}
\definecolor{gray}{rgb}{0.5,0.5,0.5}
\definecolor{mauve}{rgb}{0.58,0,0.82}
\begin{document}

\maketitleabstract

\section{Introduction}

The transformer model of \citet{vaswani} has been shown to be very successful in many different NLP tasks. The success of this model is attributed to the practice of training its large-scale variants on vast textual datasets with a language modeling objective. The models that are built in this manner are commonly termed "Large Language Models" (LLMs). This paper proposes a new task by introducing two new datasets and analyses the performance of LLMs on this task.

The task we propose is finding the locations of characters in a narrative. Spatial understanding within narrative contexts is crucial for AI systems because to truly understand a narrative, it is often necessary to understand the spatial relationships and contexts. This includes comprehending the relative positions, movements, and interactions of objects and characters. Just as humans can track characters and their changing locations while reading a story or a novel, AI models are expected to demonstrate similar capabilities. Moreover, since many linguistic constructs are deeply tied to spatial concepts, spatial understanding is important for improving language understanding in general. Prepositions, motion verbs, and even various idioms and metaphors have spatial roots. A better understanding of spatial concepts enhances the ability of a machine to comprehend and generate language.

As suggested by \citet{piper-etal-2021-narrative}, finding the locations of characters in a narrative can provide us with novel findings. For example, \citet{Wilkens} showed that the origins of US literature are not focused on New England as much as the theorists have claimed before. Also, \citet{piper-etal-2021-narrative} state that the changing social dynamics can also be observed via spatial understanding systems. As an example, one can learn about how far women and children are allowed to move or travel, either from their personal life stories, real-world news accounts, or fictional mythologies. That is why the spatial understanding task is helpful and important.

\begin{figure}[b!]
\begin{adjustbox}{width=\columnwidth}
\fbox{%
    \parbox{8.5cm}{%
        Answer the question depending on the context. \\
        Context: \textit{$<$passage extracted from a narrative$>$}; \\
        Question: \textit{$<$where is character X?$>$}; \\
        Answer: 
    }%
}
\end{adjustbox}
\caption{A prompt template for question answering.}
\label{introex}
\end{figure}

There are other works that focus on spatial understanding and propose annotated datasets, such as SpatialML \citep{spatialml}. However, our annotated datasets are specifically curated for the task of finding the locations of characters in a narrative. We manually annotated two different narrative sources: Andersen's children's stories and Persuasion book by Jane Austen. The Andersen dataset contains 249 annotations of character-location pairings, while the Persuasion dataset has 264. We prompted LLMs by providing a passage from a story or the novel and a question asking the location of a character in the passage. An example prompt template is given in Figure \ref{introex}. How the passage is extracted is explained in Section 6. In our experiments with LLMs, this task proved to be challenging. Out of the five LLMs we tested, the best-performing one for the Andersen dataset accurately identified the location in 61.85\% of the examples, while for the Persuasion dataset, the best-performing one did so in 56.06\% of the cases. The non-machine learning baseline we designed had accuracies of 55.84\% and 46.34\% for the Andersen and Persuasion datasets, respectively.

\begin{table*}[]
\centering
\begin{tabular}{|l|l|p{6cm}|l|}
\hline
\textbf{char\_no} & \textbf{character} & \textbf{location}                                                                     & \textbf{singular/plural} \\ \hline
178               & the prince         & all over the world                                                                    & singular                 \\ \hline
453               & the prince         & his palace/the prince's palace/the palace/at home/home/his home/his castle/the castle & singular                 \\ \hline
797               & the king           & the palace/outside/out/the castle/in his castle                                       & singular                 \\ \hline
850               & the princess       & outside the door/outside/out/the palace/the castle                                    & singular                 \\ \hline
1079              & the queen          & the palace/the castle/at home                                                         & singular                 \\ \hline
1173              & the queen          & the bedroom/the palace/the castle/at home                                             & singular                 \\ \hline
\end{tabular}
\caption{Annotation file for the story "The Real Princess" from the Andersen dataset.}
\label{fig:story3annotated}
\end{table*}

We also applied the in-context learning method to see if it improved the results \cite{gpt3} and saw that incorporating in-context learning improved the accuracy of this task when simple, one-sentence length examples were given in the prompt.

\section{Related Work}

\citet{booknlp1} presents a new dataset comprising literary texts annotated for six different entity categories: person, location, geo-political entity, facility, organization, and vehicle. They annotated the first 2,000 words of 100 literary books they acquired from Project Gutenberg. They annotated character and location entities, but they did not match them, unlike our work.

BABI is a dataset comprising different toy tasks to evaluate reading comprehension via question answering \cite{babi}. They suggest that a machine that can understand language should be able to succeed in these tasks. The tasks require the machine to chain facts, do simple induction, deduction, etc. There are 20 tasks in the dataset. One of the tasks that is similar to our work in the BABI dataset is Task 1, in which we are given a set of sentences with characters and location mentions, and the characters change locations within the text. Then, after every three sentences, we are queried about the final location of a character. However, the texts in the tasks are artificially generated and do not appear within a natural narrative as they do in our work.

\section{Dataset Information}
\label{sec:append-how-prod}

We annotated two datasets: Andersen children's stories and the Persuasion novel. Both of the datasets are in the English language. The Andersen dataset was annotated by the first author, and the Persuasion dataset was annotated by the second author. The annotation guideline the second author had to follow is given in Appendix A. Along with the annotation guideline, the second author was also provided with an example story from the Andersen dataset and its annotation file. During the annotation process, the second author was able to ask questions they had regarding the task and get support.

\begin{table*}[]
\centering
\begin{adjustbox}{width=\textwidth}
\begin{tabular}{cccccccccccccccc}
\textbf{Story No. }            & 1  & 2  & 3 & 5  & 7 & 8  & 9  & 10 & 11 & 12 & 13 & 15 & 16 & 17 & 18\\ \hline
\textbf{\begin{tabular}[c]{@{}c@{}}Number of\\ Annotations\end{tabular}} & 15 & 18 & 6 & 24 & 6 & 31 & 14 & 22 & 4  & 16 & 10 & 10 & 24 & 10 & 39 \\ \hline
\textbf{\begin{tabular}[c]{@{}c@{}}Number of\\ Tokens\end{tabular}} & 2558 & 2271 & 547 & 4628 & 987 & 4346 & 2760 & 4229  & 1846 & 2802 & 1301 & 1366 & 2762 & 1198 & 3015 \\ 
\end{tabular}
\end{adjustbox}
\caption{Number of annotations created and the number of tokens for each story in the Andersen dataset.}
\label{andersenstats}
\end{table*}

We acquired Andersen's Fairy Tales by H. C. Andersen book from the Project Gutenberg and selected 15 stories from it to annotate. The whole annotation was done manually. We read the story with one central question in our mind: If one is a grade-schooler reading this story, how would one answer the question of "Where is character A?" at each point in the story? When a location of a character is mentioned, the character, the location, and the place where the information mentioned in the story is written down in an annotation file. An example annotation file is given in Table \ref{fig:story3annotated}. The annotation files are in “tab-separated value” format, meaning that the columns are separated by a tab value in each line. As the table shows, there are four columns:

\begin{figure*}[ht!]
\centering
\fbox{%
    \parbox{13cm}{%
    \begingroup
    \fontsize{10pt}{8pt}\selectfont

        (...)So the Prince was appointed “Imperial Swineherd.” He had a dirty little room close by the pigsty; and there he sat the whole day, and worked. By the evening he had made a pretty little kitchen-pot. Little bells were hung all round it; and when the pot was boiling, these bells tinkled in the most charming manner, and played the old melody,

    “Ach! du lieber Augustin,
    Alles ist weg, weg, weg!”*

    * “Ah! dear Augustine!
    All is gone, gone, gone!”

But what was still more curious, whoever held his finger in the smoke of
the kitchen-pot, immediately smelt all the dishes that were cooking on
every hearth in the city--this, you see, was something quite different
from the rose.

Now the Princess happened to walk that way; and when she heard the tune,
she stood quite still, and seemed pleased; for she could play “Lieber
Augustine”; it was the only piece she knew; and she played it with one
finger. (...)

\endgroup
    }%
}
\caption{A snippet from "The Swineherd", a story from the Andersen dataset. Here, at the end of the snippet, the location of the Princess is "by the pigsty", but it is not explicitly stated.}
\label{dataex}
\end{figure*}
\begin{itemize}
\item Char\_no: The story is thought to be one giant string. Char\_no is the string index of the character at the end of the sentence where we are sure that the character (column 2) is in that location (column 3). For example, let's say that we have the following sentence in the book: "Mr. Dawson visited the park, then he saw something strange." Here, after the word "park", we are sure that Mr. Dawson is at the park. When we stop at that point and ask "Where is Mr. Dawson?" to our AI model, we expect to get "park" as the answer. The string index of the first character after the word "park" is the char\_no.
\item Character: Every character whose location is mentioned at least once in the story is noted down.
\item Location: The location of the character. All the acceptable paraphrases of the location mentioned in the story up until the char\_no should be written down. Each alternative location phrase is separated by a “/” sign.
\item Singular/plural: If the character is singular or plural. This is important as we will form the prompts for the LLMs accordingly.
\end{itemize}
There are a total of 249 annotations of character-location matchings in the Andersen stories dataset. The annotations include 101 distinct characters and 387 unique locations. It should be noted that the characters whose locations are unknown or locations without characters are not included in these numbers since they are not annotated. The number of annotations and the number of tokens for each story in the dataset are given in Table \ref{andersenstats}.

A snippet from one of the stories in the Andersen dataset, "The Swineherd", is given in Figure \ref{dataex}. This example was given because it is one of the challenging samples from the dataset that represents our task well. At the beginning of the snippet, it is mentioned that the Prince lives in a room by the pigsty where he spends his whole day and works, and he makes a magical kitchen pot there. Then, the princess walks "that way" and hears the magic pot. Therefore, we understand that the Princess is near the room of the Prince, near the pigsty. However, it is not explicitly stated by the text that the princess is near the pigsty, so understanding the correct location for that character can be difficult for an LLM.

We annotated the Persuasion dataset in the same way. Persuasion is a novel written by Jane Austen and was published in 1817. It is one of Austen's six major novels and is known for its insightful portrayal of the British gentlefolk during the early 19th century. We curated a dataset from the book containing 264 annotations, mapping 103 distinct characters to 49 unique locations. Again, characters without known locations or locations without characters are not included in these numbers.

\subsection{Discussion on the Differences between the Datasets}

Andersen and Persuasion datasets have important differences. The stories in Andersen dataset are shorter, but Persuasion is a long novel. Finding the locations of characters is easier in the Andersen dataset because the sentences and the events explained are simpler and shorter, whereas, in Persuasion, a much heavier language is used. In Persuasion, in some cases, a location is mentioned, and then later in that chapter of the book, the characters in that location are revealed without mentioning the location again. Because of this reason, Persuasion requires longer contexts to be processed together to find the locations of the characters. The more challenging nature of Persuasion is shown by the measurements reported in the following chapters. The datasets are available online. \footnotemark{} \footnotetext{\href{https://github.com/batuhan-ozyurt/Location-of-Characters-in-Narratives-Andersen-and-Persuasion-Datasets}{github.com/batuhan-ozyurt/Location-of-Characters-in-Narratives-Andersen-and-Persuasion-Datasets}}

Furthermore, Andersen dataset had many stories that included characters transforming, being referred to by different names, changing their age, etc., and therefore, it is hard to keep up with the same character within the story, which made the annotation process more difficult. Another difference between the datasets is that while Persuasion describes a single realistic story that was going in specific chronological order, with few exceptions of flashbacks and moments of reminiscing, a single story in the Andersen dataset jumps between different abstract worlds and timelines, which made it difficult to follow the notion of “now” that was used to describe the current location of characters. While Persuasion used a deeper and more complicated language, Andersen’s stories, being designed for children, did not have the same depth in language as they had in the twistedness of the plot. The main characters used in the stories differed a lot as well, since Persuasion is a novel about human life and emotions, and Andersen’s stories range from anything between following the life of a tree and satiric depictions of real-life situations.

\subsection{Inter-annotator Agreement}

After the Andersen dataset was created, it was also annotated by a group of annotators, each one of them annotating a small group of stories. At the end, we had two different annotation files for each story so that we could  measure the inter-annotator agreement. We found that 165 of the annotations were common in the two sets of annotations for the Andersen dataset. The total number of disagreements was 148, which gives the percent agreement score of 52.7\%. The F1 score turned out to be 69.0 F. Since the annotators are only given an annotation guideline and an example annotation file for a different story, and they have to read the stories carefully and do the annotations manually from scratch, it is a challenging task. Moreover, considering the fact that the probability of a random agreement to happen is almost zero, we find this agreement score to be moderately good.

\section{Evaluation Metrics}

For evaluating the performance of LLMs in the task of predicting the locations of characters given a story snippet as context, we came up with two evaluation metrics: Exact matching and fuzzy matching. The string generated by the LLM will be referred to as “output”. During the implementation of both metrics, we follow the following steps:

\begin{table}[b]
\centering
\begin{tabular}{lcc}
                                         & \textbf{Exact Match} & \textbf{Fuzzy Match} \\ \hline
\multicolumn{1}{l}{\textbf{Andersen}}   & 54.11\%              & 55.84\%              \\ \hline
\multicolumn{1}{l}{\textbf{Persuasion}} & 41.46\%              & 46.34\%              \\ \hline
\end{tabular}
\label{baselines}
\caption{Baseline scores on Andersen and Persuasion datasets.}
\end{table}

\begin{enumerate}
\item Lowercase the output and the character string.
\item Remove the stopwords from the output using NLTK’s stopwords library. Stopwords are a list of words that are insignificant and therefore filtered out -"stopped"- during pre- or postprocessing in NLP tasks. Since there is no universal list of stopwords, we are using the NLTK package’s stopwords list, which has 179 stopwords. This list includes words such as a, about, below and but.
\item Remove the stopwords from the character string using nltk’s stopwords library.
\item Remove the character’s mention from the output if the character is not mentioned in the gold locations list (In the annotation file, the gold locations are in the “location” column. There can be multiple location strings for one annotation, as one location can be expressed by different alternative wordings. This location strings list for one annotation is referred to as the “gold location list” here.).
\end{enumerate}
The outputs of the LLM are usually just a location string such as “the library” or one sentence with the character such as “The man was at the library.” Assuming the output was “The man was at the library”, following the preprocessing steps for evaluation listed above, the output string will look like these:
\begin{enumerate}
\item the man was at the library
\item man library
\item man library (Character: the man → man)
\item library
\end{enumerate}
Hence, we are left with “library”.

In the final matching step, we have two options. In exact matching, if the output and one of the gold locations are exactly the same after the preprocessing steps, we say that the output is accurate. For fuzzy matching, we use the partial ratio algorithm from the fuzzywuzzy library. The partial ratio algorithm comes up with a measure of how partially similar two strings are. The algorithm declares two strings to be partially similar if they have some of the words in a shared order. The string with the shorter length is compared with the long string’s substrings of the same length. The partial ratio score takes values between 0 and 100. We decided our threshold to be 90 in our evaluations. In other words, if the partial ratio score between the output and at least one of the locations in the gold locations list is above 90, we say that the output is accurate. 

\section{Baseline Measurements}

As a very simple baseline, we find the locations of characters by measuring the distance between a character and all of the locations mentioned in the text. Here is the algorithm for this method:

\begin{enumerate}
\item Using the annotation files, we acquire the location entity names. We find all the indices of the location entities in the novel string or the story string to create the set of locations.

\item At each line in the annotation files, we go to the index given in "char\_no" in the text; and inside the piece of string from the beginning of the text until the char\_no index, we find the index of the latest mention of the character given in that annotation line. 

\item We calculate the distances between the character mention and all locations in the location set.

\item We assign the location of that character as the location that has the minimum distance to the character mention.
\end{enumerate}

The baseline measurements are given in Table 3. 18 of the annotations from the Andersen dataset and 16 of the annotations from the Persuasion dataset could not be used in this measurement because the character strings in those annotations could not be found in the text. This is because some character entities had to be expressed in a more concise way in the annotation file.

\section{Prompting Large Language Models}

In order to see the performance of open-source large language models on the datasets, we conducted experiments with five models: T0++ \citep{t0pp}, FLAN-T5-XXL \citep{flant5xxl}, GPT-J \citep{gptj}, LLaMA 2 13B chat \citep{llama2}, and Mistral 7B Instruct \citep{mistral}. We experimented with 23 different prompt templates. 22 of the templates are taken from the Appendix section of \citet{t0pp}, and the 23rd prompt is prepared by us. The prompts we used can be found in Appendix B. 

\begin{table}[]
\centering
\begin{tabular}{cc}
\textbf{Model} & \textbf{Context Length} \\ \hline
T0++           & 1024                    \\
FLAN-T5-XXL    & 512                     \\
LLaMA-2-13B-Chat & 4096             \\
Mistral-7B-Instruct & 8192 \\
GPT-J          & 2048                   
\end{tabular}
\caption{Large language models and their maximum context length.}
\label{contextsizelim}
\end{table}

We prompted LLMs by providing a passage extracted from a story or the novel and a question asking the location of a character in the passage. The answer to the question is found at the end of the passage because we select the ending point of the passage to be the char\_no in the annotation line, and as the starting point of the passage, we go back in the story or the novel as far as we can until we reach the maximum number of allowed tokens in the prompt. The maximum context lengths of the models are given in Table \ref{contextsizelim}.

\begin{table}[b]
\centering
\begin{tabular}{|c|c|c|}
\hline
                           & \textbf{\begin{tabular}[c]{@{}c@{}}Exact \\ Matching\end{tabular}}                                                 & \textbf{\begin{tabular}[c]{@{}c@{}}Fuzzy \\ Matching\end{tabular}}                                                 \\ \hline
\textbf{\begin{tabular}[c]{@{}c@{}}Standard \\ Prompt\end{tabular}}   & \textbf{\begin{tabular}[c]{@{}l@{}}46.59\%\end{tabular}} & \textbf{\begin{tabular}[c]{@{}l@{}}57.03\%\end{tabular}} \\ \hline
\textbf{\begin{tabular}[c]{@{}c@{}}Distraction \\ Added\end{tabular}} & \begin{tabular}[c]{@{}l@{}}40.16\% \end{tabular}           & \begin{tabular}[c]{@{}l@{}}52.21\%\end{tabular}         \\ \hline
\end{tabular}
\caption{The effect of adding distractions to the prompts.}
\label{distraction}
\end{table}
The results when these five models are prompted on the Andersen and Persuasion datasets are shown in Table \ref{othermodels}. The task was found to be challenging because the best-performing model for the Andersen dataset had 61.85\% fuzzy matching accuracy, and for the Persuasion dataset, the best-performing one had 56.06\% fuzzy matching accuracy. Meanwhile, the simple baseline we designed achieved 55.84\% in Andersen and 46.34\% in Persuasion, so the LLMs did not significantly exceed the baseline. We also observe that compared to Andersen, the performance of T0++ is worse on Persuasion. Another observation is that the number of parameters affected the performance profoundly. T0++, FLAN-T5-XXL, and LLaMA 2 have 11B, 11B, and 13B parameters, respectively, while GPT-J has 6B and Mistral has 7B parameters, and this difference is also reflected in the results. It should also be noted that we found the best-performing prompt template to be different for each model. In the fuzzy matching evaluation, for models T0++, FLAN-T5-XXL, LLaMA-2-13B-Chat, Mistral-7B-Instruct, and GPT-J, Template Numbers 6, 2, 1, 1, and 1 were the best templates for the Andersen dataset, and Template Numbers 13, 10, 2, 2, and 12 were the best ones for the Persuasion dataset, respectively. This suggests that finding a perfect template is challenging, and one should try different prompt templates in different situations to find the best-performing setup. The results for different templates are available in Appendix D.

\newcommand\ChangeRT[1]{\noalign{\hrule height #1}}
\begin{table*}[]
\centering
\begin{adjustbox}{width=\textwidth}
\begin{tabular}{l|cc|cc|cc|cc|cc|}
\cline{2-11}
                                          & \multicolumn{2}{c|}{\textbf{T0++}} & \multicolumn{2}{c|}{\textbf{FLAN-T5-XXL}} & \multicolumn{2}{c|}{\textbf{\begin{tabular}[c]{@{}c@{}}LLaMA-2-\\ 13B-Chat\end{tabular}}} & \multicolumn{2}{c|}{\textbf{\begin{tabular}[c]{@{}c@{}}Mistral-7B-\\ Instruct\end{tabular}}} & \multicolumn{2}{c|}{\textbf{GPT-J}} \\ \cline{2-11} 
                                          & \textbf{Exact}   & \textbf{Fuzzy}  & \textbf{Exact}      & \textbf{Fuzzy}      & \textbf{Exact}         & \textbf{Fuzzy}        & \textbf{Exact}          & \textbf{Fuzzy}          & \textbf{Exact}   & \textbf{Fuzzy}   \\ \hline
\multicolumn{1}{|l|}{\textbf{Andersen}}   & 48.19            & 57.03         & 48.59            & 56.22             & 24.50                & 61.85               & 10.44              & 48.59                 & 23.69         & 41.77          \\ \hline
\multicolumn{1}{|l|}{\textbf{Persuasion}} & 43.94            & 48.48         & 51.89             & 56.06            & 9.85                 & 37.50               & 14.77               & 26.89                & 19.32          & 29.17          \\ \hline
\end{tabular}
\end{adjustbox}
\caption{Performance of T0++, FLAN-T5-XXL, LLaMA-2-13B-Chat, Mistral-7B-Instruct, and GPT-J models on the Andersen and Persuasion datasets when prompted. The values are in percentages.}
\label{othermodels}
\end{table*}

\begin{figure*}[]
\centering
\fbox{%
    \parbox{8.5cm}{%
        Answer the question depending on the context. \\
        Context: \textit{$<$example context 1$>$}; \\
        Question: \textit{$<$example question 1$>$}; \\
        Answer: \textit{$<$example answer 1$>$} \\
        Context: \textit{$<$example context 2$>$}; \\
        Question: \textit{$<$example question 2$>$}; \\
        Answer: \textit{$<$example answer 2$>$} \\
        Context: \textit{$<$context$>$}; \\
        Question: \textit{$<$question$>$}; \\
        Answer: 
    }%
}
\caption{2-shot in-context learning prompt example}
\label{iclex1}
\end{figure*}

\section{Adding Distractions to the Textual Prompt}

We conducted experiments to see how fragile LLMs are by adding distractions to the prompt. We aimed to observe if the LLMs failed to answer correctly when a distractive sentence was added at the end of the context. The distraction sentence we picked was “John went into the kitchen.” We worked with the T0++ model. The results are reported in Table \ref{distraction}. The metrics show that adding a distraction decreases the performance of answering the question regarding the locations of characters in the stories. This result suggests that LLMs can easily be broken by distractive sentences added to their prompts.

\section{In-Context Learning}

We experimented with in-context learning to find out if it could increase the performance of Andersen dataset task. We tried different numbers of samples in the contexts. If there are "k" number of examples in the prompt, we refer to it as a "k-shot" prompt. We also experimented with different types of samples to put in the context as examples. One in-context learning prompt example is given in Figure \ref{iclex1}.

\begin{table*}[ht!]
\centering
\begin{adjustbox}{width=\textwidth}
\begin{tabular}{lcccc|cccc}
\cline{2-9}
\textbf{}        & \multicolumn{4}{c|}{\textbf{Exact Matching}}                                                                                                                                                                                                                                             & \multicolumn{4}{c}{\textbf{Fuzzy Matching}}                                                                                                                                                                                                                                              \\ \cline{2-9} 
\textbf{}        & \textbf{\begin{tabular}[c]{@{}c@{}}Mean \\ Accuracy\end{tabular}} & \textbf{\begin{tabular}[c]{@{}c@{}}Standard \\ Deviation\end{tabular}} & \textbf{\begin{tabular}[c]{@{}c@{}}Minimum \\ Accuracy\end{tabular}} & \textbf{\begin{tabular}[c]{@{}c@{}}Maximum \\ Accuracy\end{tabular}} & \textbf{\begin{tabular}[c]{@{}c@{}}Mean \\ Accuracy\end{tabular}} & \textbf{\begin{tabular}[c]{@{}c@{}}Standard \\ Deviation\end{tabular}} & \textbf{\begin{tabular}[c]{@{}c@{}}Minimum \\ Accuracy\end{tabular}} & \textbf{\begin{tabular}[c]{@{}c@{}}Maximum \\ Accuracy\end{tabular}} \\ \hline
\textbf{1-shot}  & \cellcolor[HTML]{FFFFFF}14.52\%                                   & \cellcolor[HTML]{FFFFFF}1.91\%                                         & \cellcolor[HTML]{FFFFFF}10.04\%                                      & \cellcolor[HTML]{FFFFFF}18.07\%                                      & \cellcolor[HTML]{FFFFFF}18.67\%                                   & \cellcolor[HTML]{FFFFFF}2.08\%                                         & \cellcolor[HTML]{FFFFFF}14.46\%                                      & \cellcolor[HTML]{FFFFFF}23.29\%                                      \\ \hline
\textbf{2-shot}  & \cellcolor[HTML]{FFFFFF}15.94\%                                   & \cellcolor[HTML]{FFFFFF}1.68\%                                         & \cellcolor[HTML]{FFFFFF}12.05\%                                      & \cellcolor[HTML]{FFFFFF}20.48\%                                      & \cellcolor[HTML]{FFFFFF}20.26\%                                   & \cellcolor[HTML]{FFFFFF}1.77\%                                         & \cellcolor[HTML]{FFFFFF}17.27\%                                      & \cellcolor[HTML]{FFFFFF}24.90\%                                      \\ \hline
\textbf{3-shot}  & \cellcolor[HTML]{FFFFFF}16.14\%                                   & \cellcolor[HTML]{FFFFFF}1.85\%                                         & \cellcolor[HTML]{FFFFFF}12.05\%                                      & \cellcolor[HTML]{FFFFFF}20.08\%                                      & \cellcolor[HTML]{FFFFFF}20.43\%                                   & \cellcolor[HTML]{FFFFFF}2.12\%                                         & \cellcolor[HTML]{FFFFFF}15.66\%                                      & \cellcolor[HTML]{FFFFFF}26.10\%                                      \\ \hline
\textbf{4-shot}  & \cellcolor[HTML]{FFFFFF}16.14\%                                   & \cellcolor[HTML]{FFFFFF}1.85\%                                         & \cellcolor[HTML]{FFFFFF}12.05\%                                      & \cellcolor[HTML]{FFFFFF}20.08\%                                      & \cellcolor[HTML]{FFFFFF}20.43\%                                   & \cellcolor[HTML]{FFFFFF}2.12\%                                         & \cellcolor[HTML]{FFFFFF}15.66\%                                      & \cellcolor[HTML]{FFFFFF}26.10\%                                      \\ \hline
\textbf{6-shot}  & \cellcolor[HTML]{FFFFFF}10.54\%                                   & \cellcolor[HTML]{FFFFFF}1.56\%                                         & \cellcolor[HTML]{FFFFFF}6.43\%                                       & \cellcolor[HTML]{FFFFFF}14.06\%                                      & \cellcolor[HTML]{FFFFFF}13.37\%                                   & \cellcolor[HTML]{FFFFFF}1.94\%                                         & \cellcolor[HTML]{FFFFFF}9.24\%                                       & \cellcolor[HTML]{FFFFFF}19.28\%                                      \\ \hline
\textbf{8-shot}  & \cellcolor[HTML]{FFFFFF}5.90\%                                    & \cellcolor[HTML]{FFFFFF}1.47\%                                         & \cellcolor[HTML]{FFFFFF}2.81\%                                       & \cellcolor[HTML]{FFFFFF}9.24\%                                       & \cellcolor[HTML]{FFFFFF}7.05\%                                    & \cellcolor[HTML]{FFFFFF}1.68\%                                         & \cellcolor[HTML]{FFFFFF}3.21\%                                       & \cellcolor[HTML]{FFFFFF}10.44\%                                      \\ \hline
\textbf{14-shot} & \cellcolor[HTML]{FFFFFF}0.78\%                                    & \cellcolor[HTML]{FFFFFF}0.54\%                                         & \cellcolor[HTML]{FFFFFF}0.00\%                                       & \cellcolor[HTML]{FFFFFF}2.81\%                                       & \cellcolor[HTML]{FFFFFF}0.92\%                                    & \cellcolor[HTML]{FFFFFF}0.58\%                                         & \cellcolor[HTML]{FFFFFF}0.00\%                                       & \cellcolor[HTML]{FFFFFF}2.81\%                                       \\ \hline
\end{tabular}
\label{icl1averaged}
\end{adjustbox}
\caption{In-context-learning results on Andersen dataset. In-context examples are randomly sampled from the same dataset. The setup was run 50 times, and the average accuracies of these 50 runs, along with the standard deviation, minimum, and maximum values for the runs are given.}
\end{table*}

\begin{table*}[]
\begin{adjustbox}{width=\textwidth}
\begin{tabular}{lcccc|cccc}
\cline{2-9}
\textbf{}       & \multicolumn{4}{c|}{\textbf{Exact Matching}}                                                                                                                                                                                                                                             & \multicolumn{4}{c}{\textbf{Fuzzy Matching}}                                                                                                                                                                                                                                              \\ \cline{2-9} 
\textbf{}       & \textbf{\begin{tabular}[c]{@{}c@{}}Mean \\ Accuracy\end{tabular}} & \textbf{\begin{tabular}[c]{@{}c@{}}Standard \\ Deviation\end{tabular}} & \textbf{\begin{tabular}[c]{@{}c@{}}Minimum \\ Accuracy\end{tabular}} & \textbf{\begin{tabular}[c]{@{}c@{}}Maximum \\ Accuracy\end{tabular}} & \textbf{\begin{tabular}[c]{@{}c@{}}Mean \\ Accuracy\end{tabular}} & \textbf{\begin{tabular}[c]{@{}c@{}}Standard \\ Deviation\end{tabular}} & \textbf{\begin{tabular}[c]{@{}c@{}}Minimum \\ Accuracy\end{tabular}} & \textbf{\begin{tabular}[c]{@{}c@{}}Maximum \\ Accuracy\end{tabular}} \\ \hline
\textbf{1-shot} & 24.61\%                                                           & 1.68\%                                                                 & 19.68\%                                                              & 27.31\%                                                              & 31.28\%                                                           & 2.28\%                                                                 & 26.51\%                                                              & 36.55\%                                                              \\ \hline
\textbf{2-shot} & 35.68\%                                                           & 1.78\%                                                                 & 32.53\%                                                              & 39.36\%                                                              & 43.82\%                                                           & 2.13\%                                                                 & 38.15\%                                                              & 47.79\%                                                              \\ \hline
\textbf{3-shot} & 42.51\%                                                           & 1.71\%                                                                 & 40.16\%                                                              & 46.18\%                                                              & 51.88\%                                                           & 1.99\%                                                                 & 47.79\%                                                              & 55.82\%                                                              \\ \hline
\textbf{4-shot} & 42.51\%                                                           & 1.71\%                                                                 & 40.16\%                                                              & 46.18\%                                                              & 51.88\%                                                           & 1.99\%                                                                 & 47.79\%                                                              & 55.82\%                                                              \\ \hline
\textbf{6-shot} & 49.27\%                                                           & 1.69\%                                                                 & 44.18\%                                                              & 52.21\%                                                              & 59.20\%                                                           & 2.18\%                                                                 & 55.02\%                                                              & 63.86\%                                                              \\ \hline
\textbf{8-shot} & 49.02\%                                                           & 1.61\%                                                                 & 44.98\%                                                              & 51.81\%                                                              & 59.48\%                                                           & 1.58\%                                                                 & 56.22\%                                                              & 63.05\%                                                              \\ \hline
\end{tabular}
\end{adjustbox}
\label{icl2averaged}
\caption{In-context-learning results on Andersen dataset. In-context examples are randomly chosen from simple, one-sentence length contexts. The setup was run 50 times, and the average accuracies of these 50 runs, along with the standard deviation, minimum, and maximum values for the runs are given.}
\end{table*}

In the first set of experiments, we use question-answer pairs from the Andersen stories dataset itself. All the examples and the context of the actual question have the same number of tokens. The important note here is that the samples and the questions should all come from different stories in the dataset in order to avoid repetition in the prompt. The goal is to show examples of similar question-answer pairs to help LLM understand the task better and improve accuracy.

In the second set of experiments, the few-shot examples are simple, artificial, one-sentence-length texts such as "The beggar was begging for money in the street."  The whole list of examples is given in Appendix C. We randomly select from this list of examples and construct the prompt. The goal here is that the LLM learns about the task of matching characters with locations from simple example sentences.

The results are given in Tables 7 and 8 for the first and second sets of experiments, respectively. We conducted all the measurements using Prompt Template 1. For the zero-shot case, T0++ has an exact matching accuracy of 38.55\% and a fuzzy matching accuracy of 47.39\% with Template 1. Each setting was run 50 times, and the average accuracies for these runs, along with the standard deviation, minimum, and maximum values, are given in the tables. In the first set, the results are much lower than the zero-shot setting. Also, there is not a monotonic relationship between the number of shots and the accuracies. The results from the first set of experiments show that in-context learning does not help our task. However, in the second set of experiments, as the number of shots increases, we see an increase in performance, and after the number of examples is 3 or more, the performance is better than the zero-shot setting. We speculate that the reason why the first set does not increase the accuracy while the second setting does is the context length difference. In the first setting, the passage that the question is about has a length of approximately \begin{math}1024 / (k+1)\end{math} tokens since the context length of T0++ is 1024. This corresponds to around 114 tokens in 8-shot learning and 68 tokens in 14-shot learning, which might be too short for the model to reason about. However, in the second setting, each in-context example has an average length of 11 tokens. In 8-shot learning, this corresponds to a context length of more than 900 tokens, so the performance is not expected to be affected by the shortening of the context length significantly. However, we need to conduct further experiments to validate this hypothesis.

\section{Conclusion}

In this paper, we presented two new datasets, Andersen and Persuasion, in which we manually annotated the locations of characters in the text. We evaluated the performance of LLMs on answering questions about the character-location relations in narratives using the dataset. In our experiments with five different LLMs on these datasets formatted in a question-answering style, this task proved to be challenging because the maximum accuracy we acquired was 61.85\% for Andersen and 56.06\% for Persuasion. Furthermore, we observed that the introduction of a distracting sentence featuring an irrelevant character and location at the end of the stories reduced the accuracy scores. Our experiments with the in-context learning approach yielded varied outcomes: While using in-context examples from the same dataset worsened performance, leveraging artificial, one-sentence length examples enhanced the results.

\nocite{*}
\section{Bibliographical References}\label{sec:reference}

\bibliographystyle{lrec-coling2024-natbib}
\bibliography{lrec-coling2024-example}

\begin{thebibliography}{12}
\expandafter\ifx\csname natexlab\endcsname\relax\def\natexlab#1{#1}\fi

\bibitem[{Bamman et~al.(2019)Bamman, Popat, and Shen}]{booknlp1}
David Bamman, Sejal Popat, and Sheng Shen. 2019.
\newblock \href {https://doi.org/10.18653/v1/N19-1220} {An annotated dataset of literary entities}.
\newblock In \emph{Proceedings of the 2019 Conference of the North {A}merican Chapter of the Association for Computational Linguistics: Human Language Technologies, Volume 1 (Long and Short Papers)}, pages 2138--2144, Minneapolis, Minnesota. Association for Computational Linguistics.

\bibitem[{Brown et~al.(2020)Brown, Mann, Ryder, Subbiah, Kaplan, Dhariwal, Neelakantan, Shyam, Sastry, Askell, Agarwal, Herbert-Voss, Krueger, Henighan, Child, Ramesh, Ziegler, Wu, Winter, Hesse, Chen, Sigler, Litwin, Gray, Chess, Clark, Berner, McCandlish, Radford, Sutskever, and Amodei}]{gpt3}
Tom Brown, Benjamin Mann, Nick Ryder, Melanie Subbiah, Jared~D Kaplan, Prafulla Dhariwal, Arvind Neelakantan, Pranav Shyam, Girish Sastry, Amanda Askell, Sandhini Agarwal, Ariel Herbert-Voss, Gretchen Krueger, Tom Henighan, Rewon Child, Aditya Ramesh, Daniel Ziegler, Jeffrey Wu, Clemens Winter, Chris Hesse, Mark Chen, Eric Sigler, Mateusz Litwin, Scott Gray, Benjamin Chess, Jack Clark, Christopher Berner, Sam McCandlish, Alec Radford, Ilya Sutskever, and Dario Amodei. 2020.
\newblock \href {https://proceedings.neurips.cc/paper_files/paper/2020/file/1457c0d6bfcb4967418bfb8ac142f64a-Paper.pdf} {Language models are few-shot learners}.
\newblock In \emph{Advances in Neural Information Processing Systems}, volume~33, pages 1877--1901. Curran Associates, Inc.

\bibitem[{Chung et~al.(2022)Chung, Hou, Longpre, Zoph, Tay, Fedus, Li, Wang, Dehghani, Brahma et~al.}]{flant5xxl}
Hyung~Won Chung, Le~Hou, Shayne Longpre, Barret Zoph, Yi~Tay, William Fedus, Eric Li, Xuezhi Wang, Mostafa Dehghani, Siddhartha Brahma, et~al. 2022.
\newblock Scaling instruction-finetuned language models.
\newblock \emph{arXiv preprint arXiv:2210.11416}.

\bibitem[{Jiang et~al.(2023)Jiang, Sablayrolles, Mensch, Bamford, Chaplot, Casas, Bressand, Lengyel, Lample, Saulnier et~al.}]{mistral}
Albert~Q Jiang, Alexandre Sablayrolles, Arthur Mensch, Chris Bamford, Devendra~Singh Chaplot, Diego de~las Casas, Florian Bressand, Gianna Lengyel, Guillaume Lample, Lucile Saulnier, et~al. 2023.
\newblock Mistral 7b.
\newblock \emph{arXiv preprint arXiv:2310.06825}.

\bibitem[{Mani et~al.(2010)Mani, Doran, Harris, Hitzeman, Quimby, Richer, Wellner, Mardis, and Clancy}]{spatialml}
Inderjeet Mani, Christy Doran, Dave Harris, Janet Hitzeman, Rob Quimby, Justin Richer, Ben Wellner, Scott Mardis, and Seamus Clancy. 2010.
\newblock Spatialml: annotation scheme, resources, and evaluation.
\newblock \emph{Language Resources and Evaluation}, 44:263--280.

\bibitem[{Piper et~al.(2021)Piper, So, and Bamman}]{piper-etal-2021-narrative}
Andrew Piper, Richard~Jean So, and David Bamman. 2021.
\newblock \href {https://doi.org/10.18653/v1/2021.emnlp-main.26} {Narrative theory for computational narrative understanding}.
\newblock In \emph{Proceedings of the 2021 Conference on Empirical Methods in Natural Language Processing}, pages 298--311, Online and Punta Cana, Dominican Republic. Association for Computational Linguistics.

\bibitem[{Sanh et~al.(2022)Sanh, Webson, Raffel, Bach, Sutawika, Alyafeai, Chaffin, Stiegler, Raja, Dey, Bari, Xu, Thakker, Sharma, Szczechla, Kim, Chhablani, Nayak, Datta, Chang, Jiang, Wang, Manica, Shen, Yong, Pandey, Bawden, Wang, Neeraj, Rozen, Sharma, Santilli, Fevry, Fries, Teehan, Scao, Biderman, Gao, Wolf, and Rush}]{t0pp}
Victor Sanh, Albert Webson, Colin Raffel, Stephen Bach, Lintang Sutawika, Zaid Alyafeai, Antoine Chaffin, Arnaud Stiegler, Arun Raja, Manan Dey, M~Saiful Bari, Canwen Xu, Urmish Thakker, Shanya~Sharma Sharma, Eliza Szczechla, Taewoon Kim, Gunjan Chhablani, Nihal Nayak, Debajyoti Datta, Jonathan Chang, Mike Tian-Jian Jiang, Han Wang, Matteo Manica, Sheng Shen, Zheng~Xin Yong, Harshit Pandey, Rachel Bawden, Thomas Wang, Trishala Neeraj, Jos Rozen, Abheesht Sharma, Andrea Santilli, Thibault Fevry, Jason~Alan Fries, Ryan Teehan, Teven~Le Scao, Stella Biderman, Leo Gao, Thomas Wolf, and Alexander~M Rush. 2022.
\newblock \href {https://openreview.net/forum?id=9Vrb9D0WI4} {Multitask prompted training enables zero-shot task generalization}.
\newblock In \emph{International Conference on Learning Representations}.

\bibitem[{Touvron et~al.(2023)Touvron, Martin, Stone, Albert, Almahairi, Babaei, Bashlykov, Batra, Bhargava, Bhosale et~al.}]{llama2}
Hugo Touvron, Louis Martin, Kevin Stone, Peter Albert, Amjad Almahairi, Yasmine Babaei, Nikolay Bashlykov, Soumya Batra, Prajjwal Bhargava, Shruti Bhosale, et~al. 2023.
\newblock Llama 2: Open foundation and fine-tuned chat models.
\newblock \emph{arXiv preprint arXiv:2307.09288}.

\bibitem[{Vaswani et~al.(2017)Vaswani, Shazeer, Parmar, Uszkoreit, Jones, Gomez, Kaiser, and Polosukhin}]{vaswani}
Ashish Vaswani, Noam Shazeer, Niki Parmar, Jakob Uszkoreit, Llion Jones, Aidan~N Gomez, \L~ukasz Kaiser, and Illia Polosukhin. 2017.
\newblock \href {https://proceedings.neurips.cc/paper_files/paper/2017/file/3f5ee243547dee91fbd053c1c4a845aa-Paper.pdf} {Attention is all you need}.
\newblock In \emph{Advances in Neural Information Processing Systems}, volume~30. Curran Associates, Inc.

\bibitem[{Wang and Komatsuzaki(2021)}]{gptj}
Ben Wang and Aran Komatsuzaki. 2021.
\newblock {GPT-J-6B: A 6 Billion Parameter Autoregressive Language Model}.
\newblock \url{https://github.com/kingoflolz/mesh-transformer-jax}.

\bibitem[{Weston et~al.(2015)Weston, Bordes, Chopra, Rush, Van~Merri{\"e}nboer, Joulin, and Mikolov}]{babi}
Jason Weston, Antoine Bordes, Sumit Chopra, Alexander~M Rush, Bart Van~Merri{\"e}nboer, Armand Joulin, and Tomas Mikolov. 2015.
\newblock Towards ai-complete question answering: A set of prerequisite toy tasks.
\newblock \emph{arXiv preprint arXiv:1502.05698}.

\bibitem[{Wilkens(2013)}]{Wilkens}
Matthew Wilkens. 2013.
\newblock \href {http://www.jstor.org/stable/43817603} {The geographic imagination of civil war-era american fiction}.
\newblock \emph{American Literary History}, 25(4):803--840.

\end{thebibliography}

\newpage
\appendix
\onecolumn
\section{Annotation Guideline}

What you are going to do is to write down the locations of characters into a tab-separated values (.tsv) file. The first line of this file, that is, the header, is like this:

\textit{char\_no $\>$$\>$$\>$ character  $\>$$\>$$\>$  location    $\>$$\>$$\>$ singular/plural}

There are four columns in this file: Char\_no, character, location, and singular/plural. Think of the book as one giant string. Char\_no, is the index of character at the end of the sentence where I am sure that the character (column 2) is in that location (column 3). For example, let's say that we have the following sentence in the book:

\textit{Mr. Dawson visited the park, then he saw something strange.}

Here, after the word "park", we are sure that Mr. Dawson is at the park. When we stop at that point and ask "Where is Mr. Dawson?" to our AI model, we expect to get "park" as the answer. That is why we find char\_no like this: 
\begin{lstlisting}
path = path_to_book.txt
f = open(path, "r")
book = f.read() 
f.close()
x = book.find("then he saw something strange.")
\end{lstlisting}
You will take the 5-10 words that come after "park" and find them in the string named "book" using the "find" method. But there is a case in which you should be careful:

\textit{The park that Mr. Dawson visited was beautiful. The children were playing.}

Here, we know for sure that Mr. Dawson is at the park after the word "visited," but "The park that Mr. Dawson visited" is not a meaningful sentence. That is why we should find the index of the point in the book where the phrase "was beautiful" ends.
\begin{lstlisting}
x = story.find("The children were playing.")
\end{lstlisting}
After writing down the char\_no and character, you should write the location. You can write alternatives to the location, each one separated by a "/" sign. It should also be noted that the order of all of the locations in a single annotation line is not important, the only thing that is important is that the first location has to be the most recent. All the other location phrases can be in random order.

\newpage

\section{Prompt Templates}

1. 
Answer the question depending on the context. \\
Context: \{\{context\}\}; \\
Question: Where is \{\{character\}\}?; \\
Answer: \\

2.
What is the answer? \\
Context: \{\{context\}\}; \\
Question: Where is \{\{character\}\}?; \\
Answer: \\

3. 
\{\{context\}\} Can you tell me where \{\{character\}\} is? \\

4.
\{\{context\}\} Please tell me where \{\{character\}\} is. \\

5.
\{\{context\}\} Tell me where \{\{character\}\} is. \\

6.
\{\{context\}\} From the passage, where is \{\{character\}\}? \\

7.
\{\{context\}\} I want to know where \{\{character\}\} is. \\

8.
\{\{context\}\} I want to ask where \{\{character\}\} is. \\

9.
\{\{context\}\} What is the answer to: Where is \{\{character\}\}? \\

10.
\{\{context\}\} Find the answer to: Where is \{\{character\}\}? \\

11.
\{\{context\}\} Answer: Where is \{\{character\}\}? \\

12.
Answer the question depending on the context. \\
Context: \{\{context\}\}; \\
Question: Where is \{\{character\}\}?; \\
If you can't find the answer, please respond "unanswerable". \\
Answer:  \\

13.
What is the answer? \\
Context: \{\{context\}\}; \\
Question: Where is \{\{character\}\}?; \\
If you can't find the answer, please respond "unanswerable". \\
Answer:  \\

14.
\{\{context\}\} Can you tell me where \{\{character\}\} is? If you can't find the answer, please respond "unanswerable". \\

15.
\{\{context\}\} Please tell me where \{\{character\}\} is. If you can't find the answer, please respond "unanswerable". \\

16.
\{\{context\}\} Tell me where \{\{character\}\} is. If you can't find the answer, please respond "unanswerable". \\

17.
\{\{context\}\} From the passage, where is \{\{character\}\}? If you can't find the answer, please respond "unanswerable". \\

18.
\{\{context\}\} I want to know where \{\{character\}\} is. If you can't find the answer, please respond "unanswerable". \\

19.
\{\{context\}\} I want to ask where \{\{character\}\} is. If you can't find the answer, please respond "unanswerable". \\

20.
\{\{context\}\} What is the answer to: Where is \{\{character\}\}? If you can't find the answer, please respond "unanswerable". \\

21.
\{\{context\}\} Find the answer to: Where is \{\{character\}\}? If you can't find the answer, please respond "unanswerable". \\

22.
\{\{context\}\} Answer: Where is \{\{character\}\}? If you can't find the answer, please respond "unanswerable". \\

23.
Where is \{\{character\}\} in the following text: \{\{context\}\} Answer:

\newpage
\section{Examples For In-Context Learning}

The examples for the in-context learning task are chosen randomly from the list below. We convert them into example question-answer pairs using the templates from Appendix B. \\

1. Mary moved to the bathroom.

2. John went to the hallway.

3. Daniel travelled to the office.

4. Lisa was running in the park when she came across the two women.

5. The salesman was very happy to see the old man in his store.

6. The lady went into the room with three windows.

7. The king and the queen was walking in the courtyard.

8. The beggar was begging for money in the street.

9. Justin visited Stockholm that month.

10. The music group had another concert at the town center and the guys were there.

11. Eric journeyed to the library.

12. The priest was at the church.

\newpage
\section{Prompting Large Language Models on Andersen and Persuasion Datasets}

\begin{table}[H]
\begin{adjustbox}{width=\textwidth}
\begin{tabular}{c|cc|cc|cc|cc|cc}
                                                           & \multicolumn{2}{c|}{\textbf{T0++}}                                                                               & \multicolumn{2}{c|}{\textbf{Flan-T5-XXL}}                                                                          & \multicolumn{2}{c|}{\textbf{LLaMA 2-Chat 13B}}                                                                   & \multicolumn{2}{c|}{\textbf{Mistral}}                                                                            & \multicolumn{2}{c}{\textbf{GPT-J}}                                                                              \\ \hline   
\begin{tabular}[c]{@{}c@{}}Template \\ Number\end{tabular} & \begin{tabular}[c]{@{}c@{}}Exact \\ Match\end{tabular} & \begin{tabular}[c]{@{}c@{}}Fuzzy \\ Match\end{tabular} & \begin{tabular}[c]{@{}c@{}}Exact \\ Match\end{tabular} & \begin{tabular}[c]{@{}c@{}}Fuzzy \\ Match\end{tabular} & \begin{tabular}[c]{@{}c@{}}Exact \\ Match\end{tabular} & \begin{tabular}[c]{@{}c@{}}Fuzzy \\ Match\end{tabular} & \begin{tabular}[c]{@{}c@{}}Exact \\ Match\end{tabular} & \begin{tabular}[c]{@{}c@{}}Fuzzy \\ Match\end{tabular} & \begin{tabular}[c]{@{}c@{}}Exact \\ Match\end{tabular} & \begin{tabular}[c]{@{}c@{}}Fuzzy \\ Match\end{tabular} \\ \hline   
1                                                          & 38.96\%                                                & 47.79\%                                                & 24.90\%                                                & 28.11\%                                                & 19.28\%                                                & \textbf{61.85\%}                                       & \textbf{10.44\%}                                       & \textbf{48.59\%}                                       & \textbf{23.69\%}                                       & \textbf{41.77\%}                                       \\
2                                                          & 36.95\%                                                & 46.99\%                                                & \textbf{48.59\%}                                       & \textbf{56.22\%}                                       & \textbf{24.50\%}                                       & 59.04\%                                                & 10.44\%                                                & 37.75\%                                                & 13.65\%                                                & 25.30\%                                                \\
3                                                          & 45.78\%                                                & 56.22\%                                                & 41.77\%                                                & 49.00\%                                                & 0.40\%                                                 & 20.48\%                                                & 2.01\%                                                 & 18.07\%                                                & 0.40\%                                                 & 12.45\%                                                \\
4                                                          & 33.33\%                                                & 50.20\%                                                & 39.36\%                                                & 49.00\%                                                & 2.81\%                                                 & 22.09\%                                                & 1.20\%                                                 & 12.05\%                                                & 0.80\%                                                 & 11.65\%                                                \\
5                                                          & 36.55\%                                                & 55.02\%                                                & 39.76\%                                                & 51.81\%                                                & 2.01\%                                                 & 28.92\%                                                & 0.80\%                                                 & 22.89\%                                                & 0.00\%                                                 & 17.27\%                                                \\
6                                                          & 46.99\%                                                & \textbf{57.03\%}                                       & 44.58\%                                                & 50.20\%                                                & 0.80\%                                                 & 27.71\%                                                & 0.00\%                                                 & 10.44\%                                                & 0.00\%                                                 & 9.24\%                                                 \\
7                                                          & 32.13\%                                                & 43.78\%                                                & 34.94\%                                                & 52.61\%                                                & 2.01\%                                                 & 11.24\%                                                & 0.00\%                                                 & 6.83\%                                                 & 0.00\%                                                 & 11.65\%                                                \\
8                                                          & 31.33\%                                                & 40.56\%                                                & 33.73\%                                                & 50.20\%                                                & 1.20\%                                                 & 14.46\%                                                & 0.00\%                                                 & 6.43\%                                                 & 0.00\%                                                 & 16.06\%                                                \\
9                                                          & 35.74\%                                                & 43.78\%                                                & 45.38\%                                                & 52.61\%                                                & 3.21\%                                                 & 46.99\%                                                & 3.21\%                                                 & 24.90\%                                                & 0.80\%                                                 & 14.86\%                                                \\
10                                                         & 44.98\%                                                & 53.01\%                                                & 45.78\%                                                & 52.21\%                                                & 3.21\%                                                 & 29.32\%                                                & 0.80\%                                                 & 30.12\%                                                & 0.80\%                                                 & 10.04\%                                                \\
11                                                         & 38.15\%                                                & 46.59\%                                                & 48.19\%                                                & 54.22\%                                                & 2.81\%                                                 & 40.16\%                                                & 0.80\%                                                 & 21.69\%                                                & 0.40\%                                                 & 12.45\%                                                \\
12                                                         & 45.78\%                                                & 55.82\%                                                & 34.14\%                                                & 37.35\%                                                & 24.10\%                                                & 55.02\%                                                & 7.63\%                                                 & 29.32\%                                                & 2.01\%                                                 & 3.61\%                                                 \\
13                                                         & \textbf{48.19\%}                                       & 56.22\%                                                & 29.72\%                                                & 33.33\%                                                & 20.88\%                                                & 50.60\%                                                & 7.63\%                                                 & 27.31\%                                                & 0.00\%                                                 & 0.40\%                                                 \\
14                                                         & 38.15\%                                                & 50.60\%                                                & 29.32\%                                                & 35.74\%                                                & 3.21\%                                                 & 9.24\%                                                 & 1.61\%                                                 & 3.21\%                                                 & 0.00\%                                                 & 4.82\%                                                 \\
15                                                         & 40.56\%                                                & 50.60\%                                                & 29.32\%                                                & 34.94\%                                                & 2.41\%                                                 & 12.05\%                                                & 3.21\%                                                 & 7.23\%                                                 & 0.00\%                                                 & 6.83\%                                                 \\
16                                                         & 39.36\%                                                & 49.40\%                                                & 28.11\%                                                & 36.14\%                                                & 2.41\%                                                 & 11.24\%                                                & 1.20\%                                                 & 6.83\%                                                 & 0.00\%                                                 & 7.63\%                                                 \\
17                                                         & 46.59\%                                                & 55.42\%                                                & 32.13\%                                                & 36.14\%                                                & 1.61\%                                                 & 9.24\%                                                 & 1.61\%                                                 & 4.02\%                                                 & 0.00\%                                                 & 3.61\%                                                 \\
18                                                         & 37.35\%                                                & 45.38\%                                                & 24.10\%                                                & 30.52\%                                                & 2.81\%                                                 & 8.43\%                                                 & 2.81\%                                                 & 7.63\%                                                 & 0.00\%                                                 & 4.82\%                                                 \\
19                                                         & 37.35\%                                                & 44.98\%                                                & 26.51\%                                                & 33.33\%                                                & 2.01\%                                                 & 6.43\%                                                 & 2.41\%                                                 & 5.62\%                                                 & 0.00\%                                                 & 5.62\%                                                 \\
20                                                         & 43.78\%                                                & 52.61\%                                                & 30.52\%                                                & 34.14\%                                                & 1.20\%                                                 & 12.05\%                                                & 2.81\%                                                 & 4.82\%                                                 & 0.00\%                                                 & 3.61\%                                                 \\
21                                                         & 44.18\%                                                & 53.01\%                                                & 30.12\%                                                & 33.73\%                                                & 2.41\%                                                 & 14.06\%                                                & 3.61\%                                                 & 9.24\%                                                 & 0.00\%                                                 & 6.43\%                                                 \\
22                                                         & 38.96\%                                                & 46.99\%                                                & 28.11\%                                                & 32.93\%                                                & 1.61\%                                                 & 10.44\%                                                & 2.01\%                                                 & 4.02\%                                                 & 0.00\%                                                 & 4.02\%                                                 \\
23                                                         & 35.74\%                                                & 44.98\%                                                & 33.33\%                                                & 40.56\%                                                & 4.02\%                                                 & 11.65\%                                                & 0.40\%                                                 & 4.42\%                                                 & 0.40\%                                                 & 3.21\%                                              \\ \hline   
\textbf{Average}                                           & \textbf{39.86\%}                                       & \textbf{49.87\%}                                       & \textbf{34.89\%}                                       & \textbf{41.96\%}                                       & \textbf{5.69\%}                                        & \textbf{24.90\%}                                       & \textbf{2.90\%}                                        & \textbf{15.37\%}                                       & \textbf{1.87\%}                                        & \textbf{10.32\%}                                      
\end{tabular}
\end{adjustbox}
\caption{Performance of T0++, FLAN-T5-XXL, LLaMA 2-Chat 13B, Mistral-7B-Instruct, and GPT-J models on the Andersen dataset when prompted with 23 different prompt templates.}
\end{table}

\begin{table}[]
\begin{adjustbox}{width=\textwidth}
\begin{tabular}{c|cc|cc|cc|cc|cc}
                                                           & \multicolumn{2}{c|}{\textbf{T0++}}                                                                               & \multicolumn{2}{c|}{\textbf{Flan-T5-XXL}}                                                                          & \multicolumn{2}{c|}{\textbf{LLaMA 2-Chat 13B}}                                                                   & \multicolumn{2}{c|}{\textbf{Mistral}}                                                                            & \multicolumn{2}{c}{\textbf{GPT-J}}                                                                              \\ \hline
\begin{tabular}[c]{@{}c@{}}Template \\ Number\end{tabular} & \begin{tabular}[c]{@{}c@{}}Exact \\ Match\end{tabular} & \begin{tabular}[c]{@{}c@{}}Fuzzy \\ Match\end{tabular} & \begin{tabular}[c]{@{}c@{}}Exact \\ Match\end{tabular} & \begin{tabular}[c]{@{}c@{}}Fuzzy \\ Match\end{tabular} & \begin{tabular}[c]{@{}c@{}}Exact \\ Match\end{tabular} & \begin{tabular}[c]{@{}c@{}}Fuzzy \\ Match\end{tabular} & \begin{tabular}[c]{@{}c@{}}Exact \\ Match\end{tabular} & \begin{tabular}[c]{@{}c@{}}Fuzzy \\ Match\end{tabular} & \begin{tabular}[c]{@{}c@{}}Exact \\ Match\end{tabular} & \begin{tabular}[c]{@{}c@{}}Fuzzy \\ Match\end{tabular} \\ \hline
1                                                          & 32.58\%                                                & 35.98\%                                                & 36.36\%                                                & 39.39\%                                                & 9.09\%                                                 & \textbf{37.50\%}                                       & \textbf{14.77\%}                                       & 25.00\%                                                & 12.50\%                                                & 22.35\%                                                \\
2                                                          & 34.09\%                                                & 37.12\%                                                & 51.14\%                                                & 55.68\%                                                & \textbf{9.85\%}                                        & \textbf{37.50\%}                                       & 14.02\%                                                & \textbf{26.89\%}                                       & 12.12\%                                                & 21.97\%                                                \\
3                                                          & 22.73\%                                                & 32.58\%                                                & 33.33\%                                                & 38.64\%                                                & 0.38\%                                                 & 1.89\%                                                 & 0.76\%                                                 & 4.55\%                                                 & 0.38\%                                                 & 4.92\%                                                 \\
4                                                          & 25.00\%                                                & 33.33\%                                                & 29.17\%                                                & 34.09\%                                                & 0.38\%                                                 & 3.41\%                                                 & 0.00\%                                                 & 5.30\%                                                 & 0.76\%                                                 & 3.79\%                                                 \\
5                                                          & 24.24\%                                                & 32.20\%                                                & 26.89\%                                                & 35.61\%                                                & 0.38\%                                                 & 3.03\%                                                 & 0.38\%                                                 & 8.33\%                                                 & 1.89\%                                                 & 4.92\%                                                 \\
6                                                          & 22.73\%                                                & 25.76\%                                                & 46.21\%                                                & 51.14\%                                                & 0.76\%                                                 & 7.95\%                                                 & 0.38\%                                                 & 4.17\%                                                 & 0.76\%                                                 & 7.95\%                                                 \\
7                                                          & 23.86\%                                                & 31.06\%                                                & 19.32\%                                                & 27.65\%                                                & 1.52\%                                                 & 5.30\%                                                 & 0.76\%                                                 & 1.52\%                                                 & 1.14\%                                                 & 6.44\%                                                 \\
8                                                          & 25.76\%                                                & 33.71\%                                                & 19.70\%                                                & 29.55\%                                                & 2.65\%                                                 & 10.61\%                                                & 1.14\%                                                 & 4.55\%                                                 & 2.27\%                                                 & 9.47\%                                                 \\
9                                                          & 25.76\%                                                & 27.27\%                                                & \textbf{51.89\%}                                       & 55.68\%                                                & 2.65\%                                                 & 12.12\%                                                & 1.14\%                                                 & 6.82\%                                                 & 0.38\%                                                 & 11.74\%                                                \\
10                                                         & 30.68\%                                                & 32.58\%                                                & \textbf{51.89\%}                                       & \textbf{56.06\%}                                       & 0.38\%                                                 & 6.44\%                                                 & 0.00\%                                                 & 2.27\%                                                 & 0.76\%                                                 & 5.30\%                                                 \\
11                                                         & 25.76\%                                                & 28.03\%                                                & 46.97\%                                                & 51.89\%                                                & 7.95\%                                                 & 25.00\%                                                & 0.76\%                                                 & 8.71\%                                                 & 1.52\%                                                 & 9.85\%                                                 \\
12                                                         & 40.53\%                                                & 44.70\%                                                & 31.06\%                                                & 35.23\%                                                & 0.76\%                                                 & 2.27\%                                                 & 2.65\%                                                 & 8.33\%                                                 & \textbf{19.32\%}                                       & \textbf{29.17\%}                                       \\
13                                                         & \textbf{43.94\%}                                       & \textbf{48.48\%}                                       & 29.55\%                                                & 33.71\%                                                & 1.14\%                                                 & 2.65\%                                                 & 2.65\%                                                 & 9.09\%                                                 & 18.18\%                                                & 26.52\%                                                \\
14                                                         & 35.23\%                                                & 38.64\%                                                & 24.62\%                                                & 27.65\%                                                & 0.00\%                                                 & 1.89\%                                                 & 0.38\%                                                 & 4.17\%                                                 & 0.38\%                                                 & 4.17\%                                                 \\
15                                                         & 34.85\%                                                & 39.02\%                                                & 26.52\%                                                & 30.30\%                                                & 0.00\%                                                 & 3.79\%                                                 & 0.00\%                                                 & 3.41\%                                                 & 0.00\%                                                 & 4.92\%                                                 \\
16                                                         & 35.98\%                                                & 40.15\%                                                & 26.89\%                                                & 29.92\%                                                & 0.38\%                                                 & 2.27\%                                                 & 0.00\%                                                 & 6.44\%                                                 & 0.38\%                                                 & 4.17\%                                                 \\
17                                                         & 35.98\%                                                & 40.53\%                                                & 30.30\%                                                & 32.58\%                                                & 0.00\%                                                 & 1.14\%                                                 & 0.00\%                                                 & 2.27\%                                                 & 0.00\%                                                 & 3.03\%                                                 \\
18                                                         & 28.41\%                                                & 34.09\%                                                & 24.24\%                                                & 28.03\%                                                & 0.00\%                                                 & 3.41\%                                                 & 0.38\%                                                 & 4.92\%                                                 & 0.38\%                                                 & 2.65\%                                                 \\
19                                                         & 22.35\%                                                & 29.55\%                                                & 21.59\%                                                & 26.14\%                                                & 0.00\%                                                 & 2.65\%                                                 & 0.00\%                                                 & 4.17\%                                                 & 0.00\%                                                 & 3.41\%                                                 \\
20                                                         & 34.85\%                                                & 36.74\%                                                & 31.82\%                                                & 35.61\%                                                & 0.38\%                                                 & 3.41\%                                                 & 0.00\%                                                 & 3.03\%                                                 & 0.38\%                                                 & 2.65\%                                                 \\
21                                                         & 40.15\%                                                & 41.67\%                                                & 30.30\%                                                & 33.71\%                                                & 0.00\%                                                 & 5.68\%                                                 & 0.00\%                                                 & 3.03\%                                                 & 0.00\%                                                 & 2.65\%                                                 \\
22                                                         & 38.64\%                                                & 42.05\%                                                & 26.89\%                                                & 30.30\%                                                & 0.00\%                                                 & 3.03\%                                                 & 0.00\%                                                 & 6.06\%                                                 & 0.00\%                                                 & 2.65\%                                                 \\
23                                                         & 22.35\%                                                & 27.27\%                                                & 35.23\%                                                & 39.77\%                                                & 0.00\%                                                 & 3.79\%                                                 & 0.38\%                                                 & 7.95\%                                                 & 0.00\%                                                 & 3.79\%                                                 \\ \hline
\textbf{Average}                                           & \textbf{30.71\%}                                       & \textbf{35.33\%}                                       & \textbf{32.69\%}                                       & \textbf{37.32\%}                                       & \textbf{1.68\%}                                        & \textbf{8.12\%}                                        & \textbf{1.76\%}                                        & \textbf{7.00\%}                                        & \textbf{3.19\%}                                        & \textbf{8.63\%}                                       
\end{tabular}
\end{adjustbox}
\caption{Performance of T0++, FLAN-T5-XXL, LLaMA 2-Chat 13B, Mistral-7B-Instruct, and GPT-J models on the Persuasion dataset when prompted with 23 different prompt templates.}
\end{table}
\end{document}